\documentclass[10 pt, conference]{ieeeconf}
\usepackage{amsfonts,amsmath}
\usepackage{amssymb} 
\usepackage{nicefrac}
\usepackage{tikz}
\usepackage{pgfplots}
\usetikzlibrary{backgrounds}
\usepackage{pgfplots}

\renewcommand{\phi}{\varphi}

\usepackage{amssymb,url,amsmath,dblfloatfix,lipsum}
\usepackage{subfigure}
\usepackage{balance}
\usepackage{array}
\usepackage{hyperref}
\usepackage[url=false,firstinits=true,backend=bibtex]{biblatex}
\IEEEoverridecommandlockouts                              
\begin{document}
\title{CapriDB - Capture, Print, Innovate: A Low-Cost Pipeline and Database \\for Reproducible Manipulation Research}

\author{Florian T. Pokorny*, Yasemin Bekiroglu*, Karl Pauwels, Judith B\"utepage, Clara Scherer, Danica Kragic
    \thanks{*These authors contributed equally to the paper.\newline
        All authors are with the Centre for Autonomous Systems,
        CAS/CVAP,\newline KTH Royal Institute of Technology, Sweden, \newline \
{\tt{\{yaseminb, fpokorny, kpauwels, butepage, cescha, dani\}@kth.se}}}}
\maketitle


\begin{abstract}
We present a novel approach and database which combines the inexpensive generation of 3D object models via
monocular or RGB-D camera images with 3D printing and a state of the art object tracking algorithm. Unlike recent
efforts towards the creation of 3D object databases for robotics, our approach does not require expensive and
controlled 3D scanning setups and enables anyone with a camera to scan, print and track complex objects for
manipulation research. The proposed approach results in highly
detailed mesh models whose 3D printed replicas are at times difficult to distinguish from the original. A key
motivation for utilizing 3D printed objects is the ability to precisely control and vary object properties such as the
mass distribution and size in the 3D printing process to obtain reproducible conditions for robotic manipulation
research. We present CapriDB - an extensible database resulting from this approach containing initially 40
textured and 3D printable mesh models together with tracking features to facilitate the adoption of the proposed
approach.
\end{abstract}

\begin{figure*}[t]
      \centering
      \includegraphics[height=0.25\textwidth]{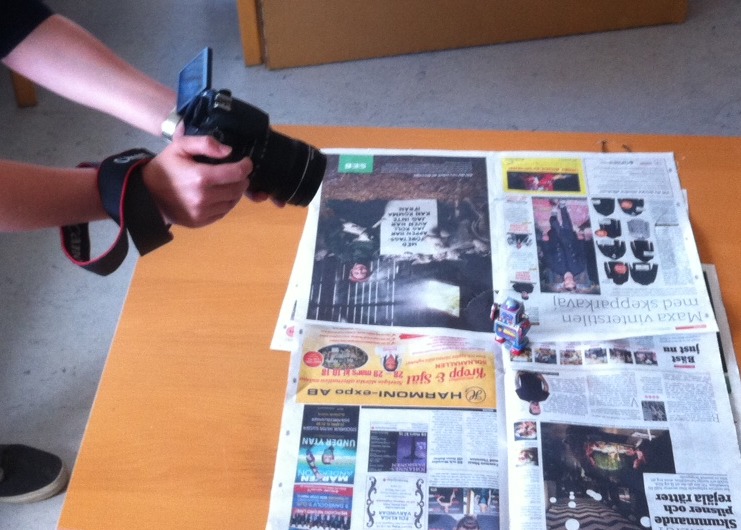}
      \includegraphics[height=0.25\textwidth]{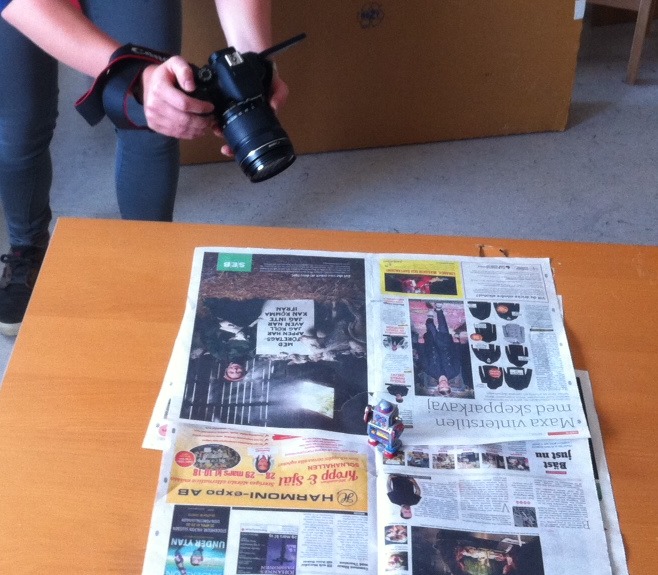}
       \includegraphics[height=0.25\textwidth]{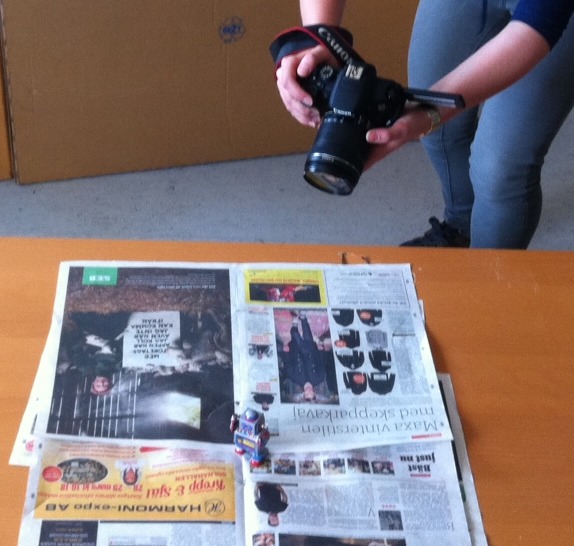}

\subfigure[Real object]{ \includegraphics[height=0.25\textwidth]{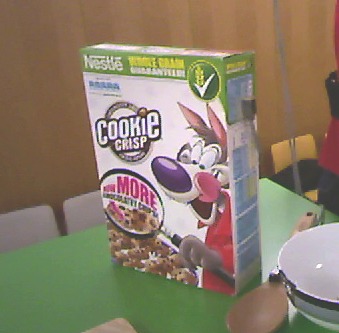}} \hspace{5pt}
      \subfigure[Printed object]{ \includegraphics[height=0.25\textwidth]{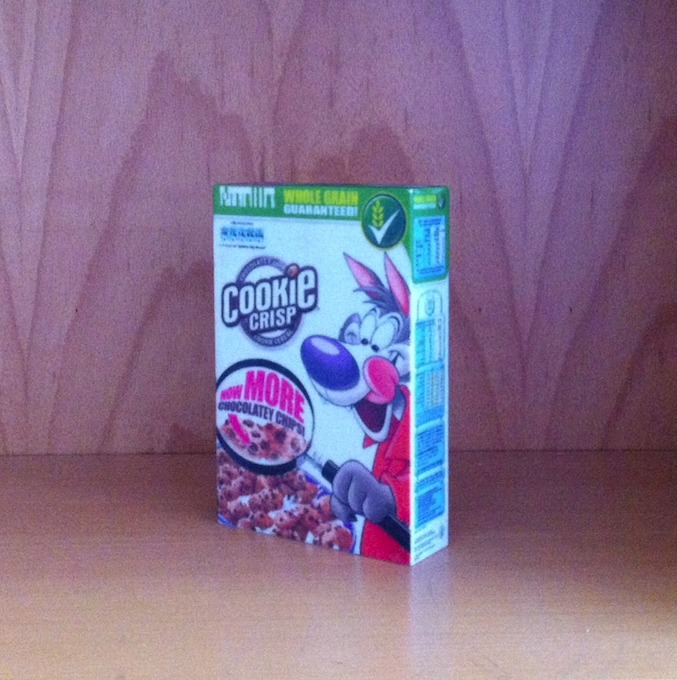}} \hspace{5pt}
      \subfigure[Tracking and grasping the printed object]{
      \includegraphics[height=0.25\textwidth]{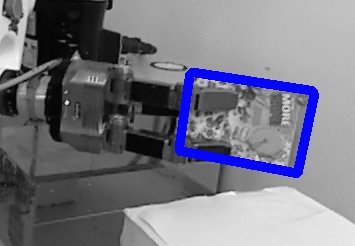}}\hspace{5pt}
      \caption{Outline of the proposed data processing approach. Upper row: Images of a small toy robot (in the center of
          figures) are captured from various arbitrary angles with a regular camera.  Lower row: An example of an acquired textured mesh-model
           model of a cerial box is 3D printed in color. The resulting 3D printed object is tracked and grasped with a robotic Schunk SDH dexterous hand.}
      \label{fig:system}
\end{figure*} 
\section{Introduction}
In this work, we focus on three fundamental steps that a robot needs to perform to manipulate an object: \emph{object
data acquisition, object representation and tracking}. During the \emph{object data acquisition phase}, sensors are used
to obtain a geometric model of the object. Typically this results in an \emph{object representation} such as a mesh
model or point-cloud which serves as input to grasping and manipulation planning algorithms such as
\parencite{Saut2012a,Borst2003a, Huebner2008a}. In the manipulation phase, the object's position and orientation needs
to be \emph{tracked} -- initially, in order to execute a planned grasp, but also during the manipulation, for example to
detect slippage. A key problem manipulation research is facing is the difficulty in reproducing results and the
complexity of benchmarking in this domain. In the last decade, a large number of groups have proposed various
benchmarking schemes, and several 3D object databases have been developed \parencite{yale, ycb, columbia, kit}. However,
these attempts have unfortunately only found partial adaptation in the research community. In our opinion, the following
have in particular inhibited widespread adoption:
\begin{itemize}
    \item Sensor dependent object models. Many works \parencite{ycb, kit} rely on costly or specially designed scanning setups.
        Our approach only requires a single hand held camera.
    \item Unavailability of objects involved: Databases such as \parencite{kit} suffer from the lack of
        worldwide availability of the objects involved. While the authors of \parencite{ycb} propose to post object
        benchmark datasets to interested researchers, our approach is to instead
        solve this problem by incorporating the use of 3D printed real world
        objects which can be printed and reproduced in a decentralized manner anywhere in the world.
    \item Size constraints: Mostly objects of size $5$ to $50cm$ have been considered in databases \parencite{yale, ycb,
        columbia, kit} due to sensor constraints, and many of these objects can be manipulated only by robotic hands of compatible size. 
        3D printing allows us to scale objects to within the 3D printer's capabilities and monocular images can be taken of 
        objects with a wide range of scales. 
    \item Dependence on material properties: 3D printing/milling, allows researchers to study the impact of object material
        properties in isolation, allowing researchers to create objects in a large number of materials and with controlled
        mass distributions.
    \item Currently, databases do not provide a reference visual tracking system, resulting in a large
        source of error and discrepancy between experimental setups.
        We incorporate our approach with the state of the art and freely available real-time visual tracking system 
        \parencite{pauwels_real-time_2013}.
\end{itemize}
To address the above issues, we introduce a 3D object database for manipulation research as well as an associated efficient and low-cost workflow 
to capture, print and track new objects. Figure~\ref{fig:system} outlines the steps of our approach. 
We utilize Autodesk's 123D catch software to acquire object models. This only requires camera images of the object taken from a variety
of arbitrary angles around the object without custom setup. 
We present details on a resulting object database containing mesh and texture information for 40 objects
as well as reference tracking image frames. Our approach is complimentary to recent efforts 
such as the recently proposed YCB database \parencite{ycb} which, unlike our work,
focuses on benchmarking protocols besides providing a set of objects scanned with a particular high-quality scanning rig. Additional key differences of our approach include the use of 3D printing rather than relying on the delivery of original objects, the
integration with a particular tracking solution as well as the low-cost extensibility of our dataset which does not rely on
a specific scanning setup or object scale but only on a hand-held camera. The database 
and associated documentation is hosted at \url{http://www.csc.kth.se/capridb/}.
In this paper, we furthermore illustrate potential applications of our approach. In particular, we verify that the
obtained object models can be 3D printed with texture and that the pose of these printed objects can be tracked
successfully. We furthermore perform initial grasping experiments using estimated poses of printed objects which are calculated using the mesh-models obtained from the original real-world objects.


\begin{figure*}[th]
    \centering
\includegraphics[width=0.40\textwidth]{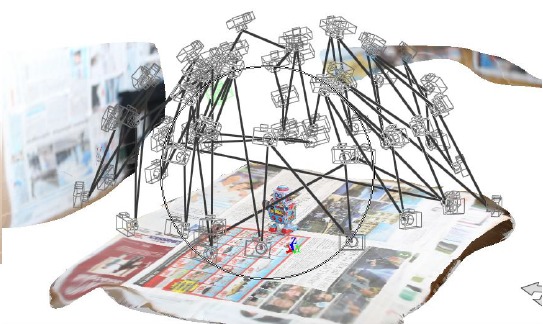}
\includegraphics[width=0.40\textwidth]{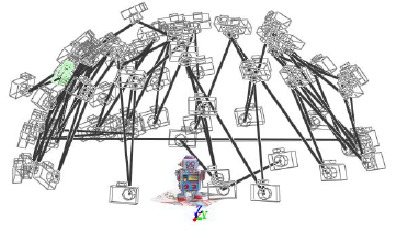}
\includegraphics[width=0.12\textwidth]{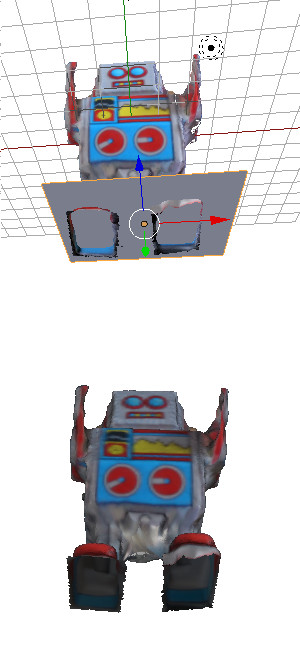}
\caption{Left and middle figures: Construction of the 3D model with Autodesk's free 123D catch application. The
reconstructed camera poses and the central object pose is displayed. Rightmost figures: Post-processing of the acquired
textured mesh model, where the mesh is made watertight and surface areas not belonging to the object are removed.}
      \label{fig:autodesk}
\end{figure*} 
\section{Methodology}
In this section, we describe the key components of our data processing pipeline: 3D model construction, 3D printing, and tracking. 
\label{sec:methodology}
\subsection{Textured 3D Model Construction}
While current grasp databases often rely on carefully calibrated 
specific capturing equipments \parencite{ycb, kit}, our approach is to use a simple digital camera
in conjunction with a freely available 3D reconstruction software to capture high-quality 3D objects.
This approach has recently become possible due to the availability of high-quality 3D reconstruction software relying
only on monocular images. To reconstruct a 3D model from a collection of photos, we utilize the web-based free Autodesk 123D catch service \parencite{autodesk}\footnote{Other solutions with similar quality are available, e.g. Agisoft PhotoScan \parencite{agisoft}.}
using approximately 40 pictures of the object from various angles.
To improve the quality of reconstruction, we place the objects on a textured background consisting of a collection of
newspapers. Figure~\ref{fig:autodesk} displays a partial screenshot of the software, illustrating the automatically
reconstructed camera positions. The scanned object is visible in the center of this visualization.

\subsection{3D Model Postprocessing}
The acquired 3D mesh model requires post-processing in order to result in a clean and fully specified
model\footnote{Detailed instructions regarding this process are available on the website.}.

Firstly, the metric dimensions of the model have to be specified in centimeters with the help of reference points for which we use
the Autodesk 123D catch software. As the initially obtained 3D mesh model contains not only the object but also some parts of the surrounding
environment, such as the surface on which the object might rest, these extraneous parts of the extracted mesh need to be
removed. We use the open source software Meshlab \parencite{meshlab} for this purpose. Figure~\ref{fig:autodesk}
illustrates post-processing steps where areas that do not belong to an object are manually removed from the initial model. In the final manual
processing step, holes in the mesh are closed. Holes arise, for example on the underside of the object, when the object rests on a planar surface when the photos are
taken. For the hole filling, we used the open source 3D modelling software Blender \parencite{blender}, which also can be
used for rotating and scaling the models as desired. Furthermore, we use a specific object pose tracker, which we 
describe later, to demonstrate that the pose of these models can be determined. The tracker requires the dimensions
of the mesh model to be provided in meters, in accordance with the ROS convention. Therefore, as a final postprocessing
step, the models are scaled accordingly. After this processing step, we
obtain a mesh model whose geometry is stored in Wavefront OBJ format, a mesh to texture mapping stored in MDL format
as well as a texture file, which is stored as a JPEG image.

\subsection{3D Printing Textured Objects}
 \begin{figure}[th]
      \centering
\includegraphics[width=0.45\textwidth]{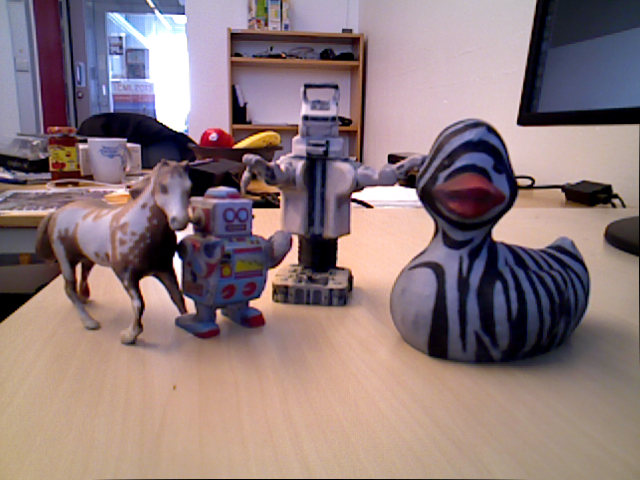}
\includegraphics[width=0.45\textwidth]{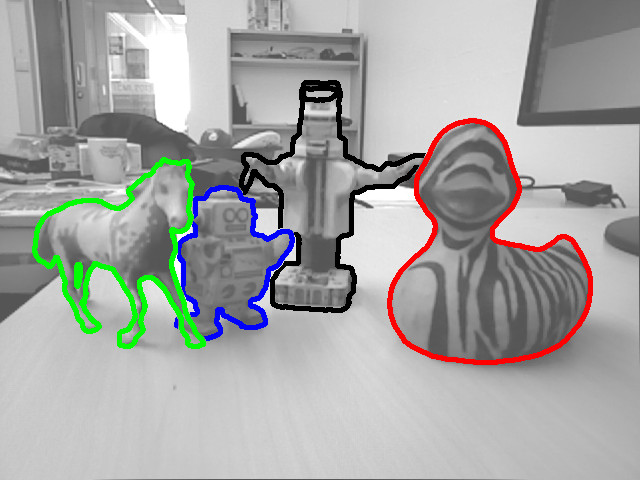}
      \caption{Top: Examples of 3D printed objects whose 3D textured model was acquired using the proposed methodology: A
      horse model, a toy robot, a rescaled PR2 robot and a toy duck. Bottom: Pose tracking results of the printed objects based 
  on textured models acquired on the originals.}
      \label{fig:tracking}
\end{figure}
 \begin{figure}[th]
      \centering
\includegraphics[width=0.45\textwidth]{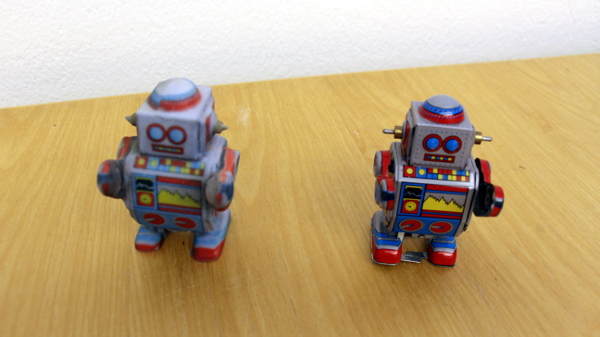}
\includegraphics[width=0.45\textwidth]{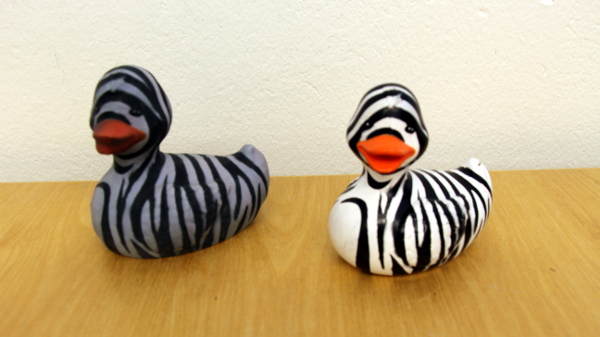}
\includegraphics[width=0.45\textwidth]{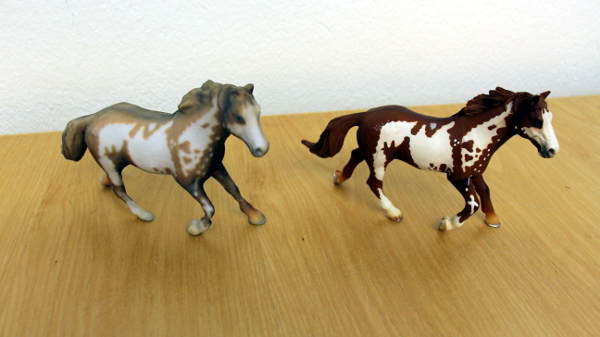}
\caption{Side by side comparison of original models (right) and 3D printed objects (left).}
      \label{fig:comparison}
\end{figure}
Our goal is to make objects accessible to everyone both as 3D mesh models and in physical/graspable forms. The rapidly
advancing field of 3D printing makes it possible to 3D print objects rather than obtaining originals. A large range of
on-line services offer to print highly textured objects in color. This allows anyone to reproduce objects based on the provided 3D mesh models and to use these for robotic manipulation research\footnote{Some of the available printing services are Shapeways (US) \parencite{shapeways}, Cubify Cloud Print (US)
\parencite{cubifycloudprint}, Sculpteo (France) \parencite{sculpteo}, iMaterialise (Belgium) \parencite{imaterialise}}. We have
printed several objects (see Figure \ref{fig:models} and Figure \ref{fig:comparison}) through the company iMaterialise \parencite{imaterialise}, see Section \ref{sec:grasping}.
Note that 3D printing also enables us to scale objects as desired, vary the internal mass distribution and select a wide
range of object materials. We believe this opens up promising new possibilities to study frictional and dynamic behavior in
robotic manipulation in a controlled fashion and independently of shape. Figure \ref{fig:tracking} displays examples of
printed objects which we scanned and printed.

\begin{figure*}[ht!]
\null\hfill
\includegraphics[height=0.2\textwidth]{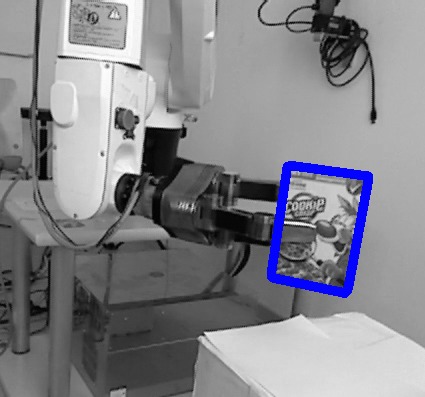}
\hfill
\includegraphics[height=0.2\textwidth]{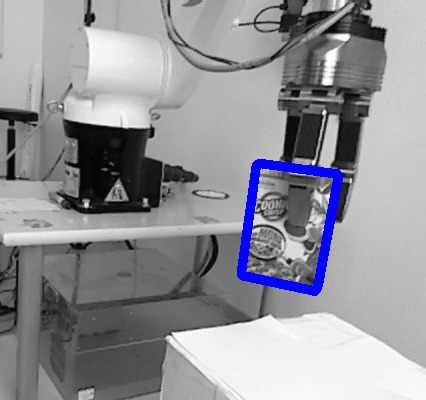}
\hfill
\includegraphics[height=0.2\textwidth]{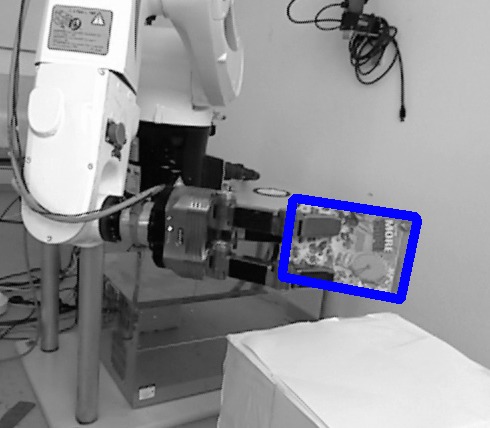}
\hfill
\includegraphics[height=0.2\textwidth]{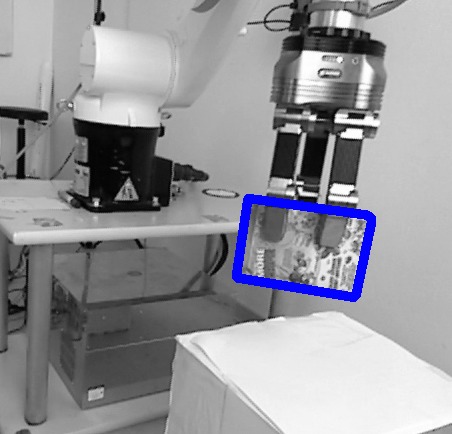}
\hfill\null
    \caption{Grasping experiments with the Kuka arm, Schunk Hand and the printed box: (a) Side grasp with the box when
    it is standing upright originally, (b) top grasp when the box is standing upright (c) side grasp when the object is
    lying sideways and the back side of the object is visible to the tracker, (d) top grasp when the object is lying
    sideways. These experiments were performed based on the object poses estimated by the tracker which used the 3D
    models obtained from the real objects, illustrating that the texture of the printed object matched the original texture sufficiently well. For tracking, images from 
    a Kinect sensor were used. The object can continuously be tracked during grasping and lifting. The blue frames around the objects indicate the tracked poses.
    }
    \label{fig:KukaSdhExp}
\end{figure*}

\subsection{Real-Time Tracking and Pose Estimation }
We use a state-of-the-art image-based object pose estimation method that uses sparse keypoints to detect, and dense
motion and depth information to track the full six degrees-of-freedom pose in real-time \parencite{pauwels_real-time_2013,
pauwels_2015}. This method achieves high accuracy and robustness by exploiting the rich appearance and shape information provided
by the models in our database. This pose estimation method is publicly available as a ROS
module\footnote{\url{www.karlpauwels.com/simtrack}}. We validate our methodology by successfully detecting and tracking
the pose of printed models on the basis of the mesh models generated from the original objects. An example tracking
result is shown in Figure \ref{fig:tracking} for a scene with occlusions and multiple 3D printed objects
and in Figure \ref{fig:pr2tracking}, where a PR2 robot's onboard arm camera is used to track several 3D printed objects.
Both RGB and RGB-D cameras can be used with this approach.


\begin{figure}[t]
      \centering
      \includegraphics[width=0.47\textwidth]{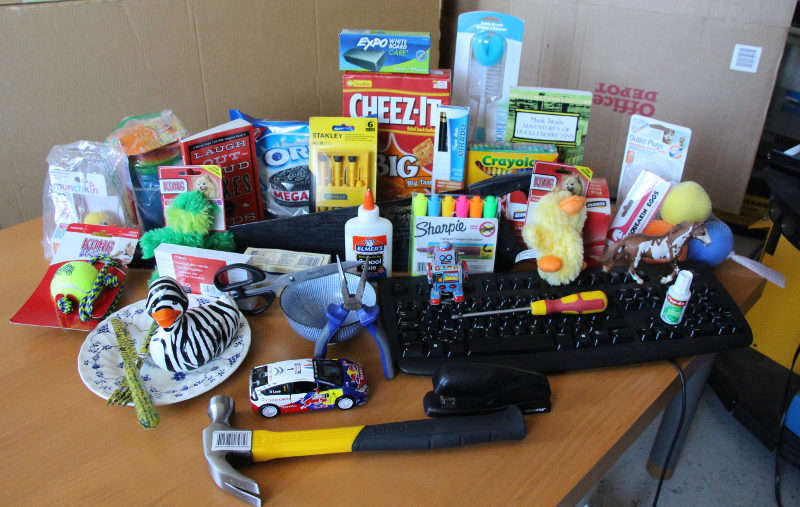}
   \caption{Initial set of 40 objects in the core database.}
      \label{fig:objects}
\end{figure} 

\subsection{Application Scenarios}
Here, we highlight various potential directions that the proposed approach and database could be used for:
\subsubsection{Integrated tracking and grasp planning}
\label{sec:grasping}
\begin{figure}[th]
\null\hfill
\includegraphics[height=0.3\textwidth]{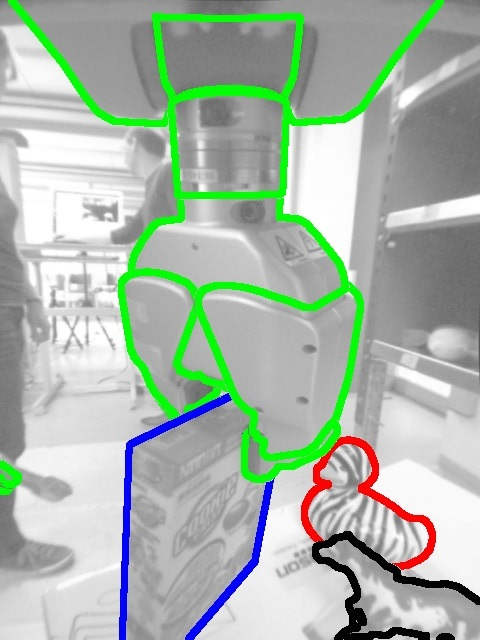}
\hfill
\includegraphics[height=0.3\textwidth]{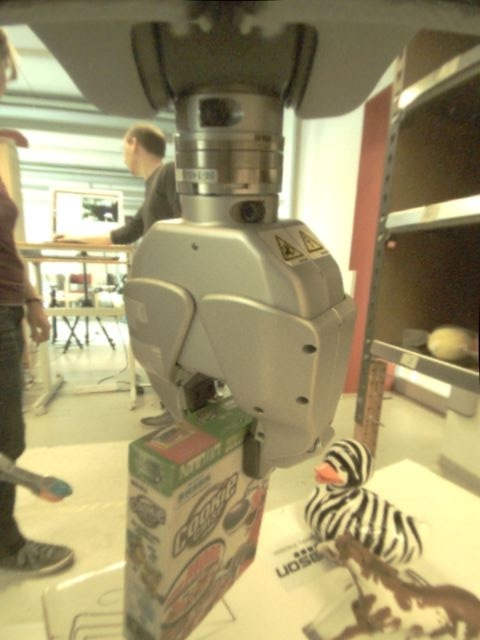}
\hfill\null
\caption{Example tracking of printed objects based on the PR2 robot's arm camera.}
\label{fig:pr2tracking}
\end{figure}
\begin{figure*}[th]
    \null\hfill
      \includegraphics[width=0.18\textwidth]{figures/boxsmall.jpg}
      \hfill
      \includegraphics[width=0.18\textwidth]{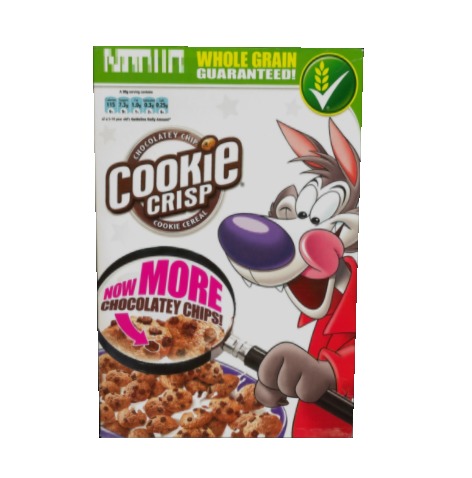}
      \hfill
      \includegraphics[width=0.18\textwidth]{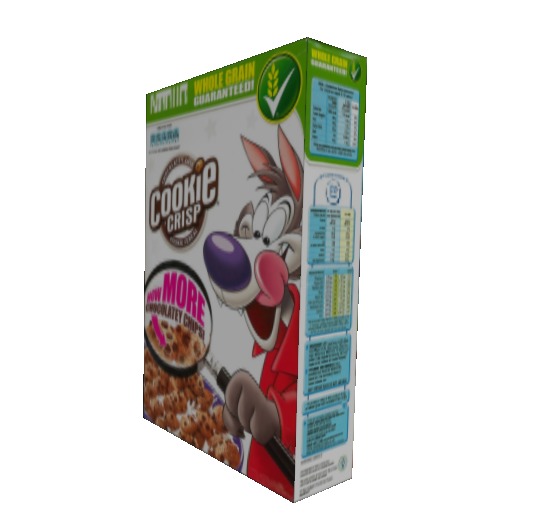}
      \hfill
      \includegraphics[width=0.18\textwidth]{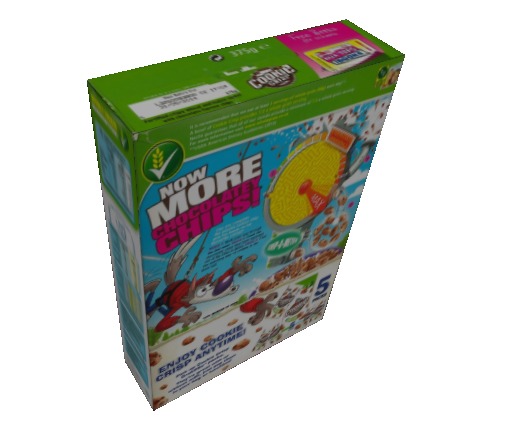}
      \hfill\null
      \vspace{10pt} 

      \null\hfill
      \includegraphics[width=0.18\textwidth]{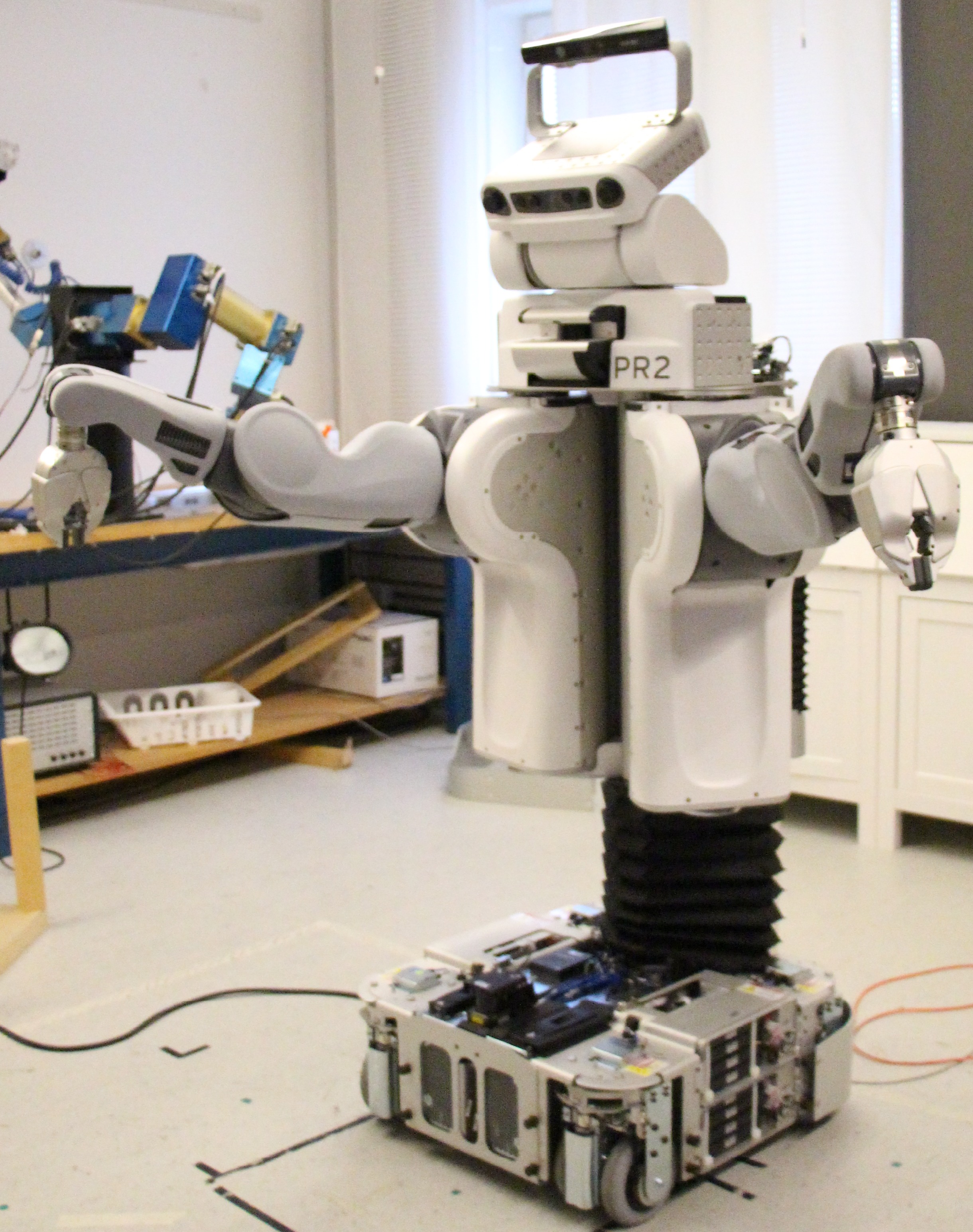}
      \hfill
      \includegraphics[width=0.2\textwidth]{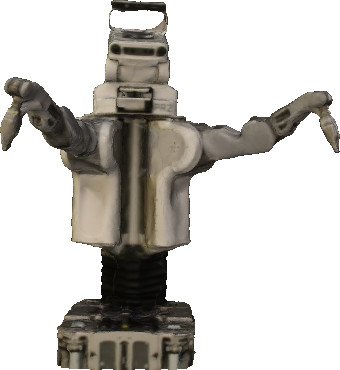} 
      \hfill
      \includegraphics[width=0.13\textwidth]{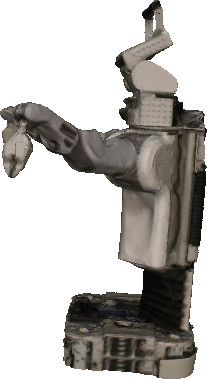} 
      \hfill
      \includegraphics[width=0.16\textwidth]{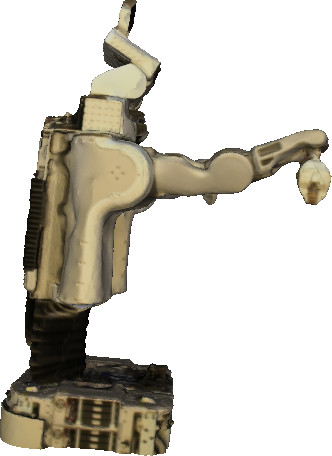}
      \null\hfill\\

      \null\hfill
 \includegraphics[width=0.2\textwidth]{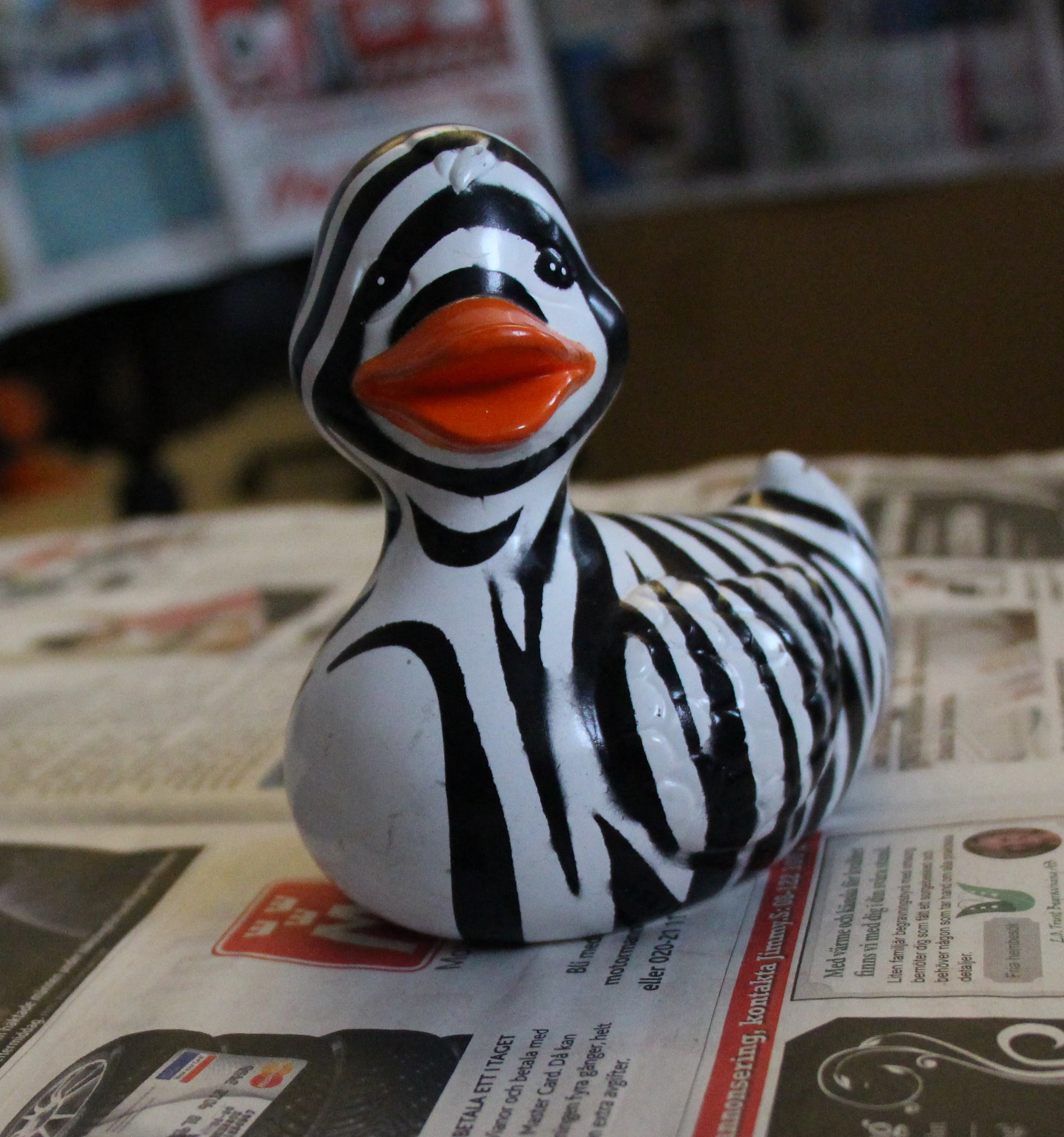} 
      \hfill
      \includegraphics[width=0.13\textwidth]{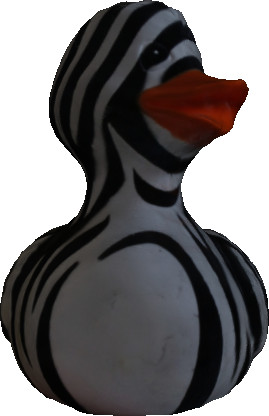} 
      \hfill
      \includegraphics[width=0.16\textwidth]{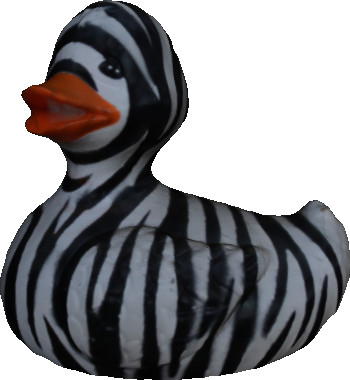} 
      \hfill
      \includegraphics[width=0.16\textwidth]{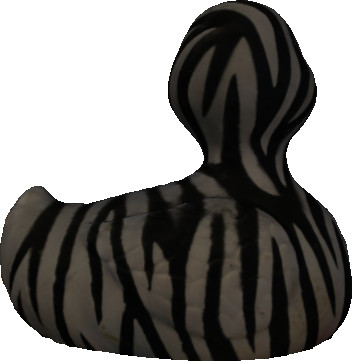}
      \null\hfill\\

      \null\hfill
      \includegraphics[width=0.17\textwidth]{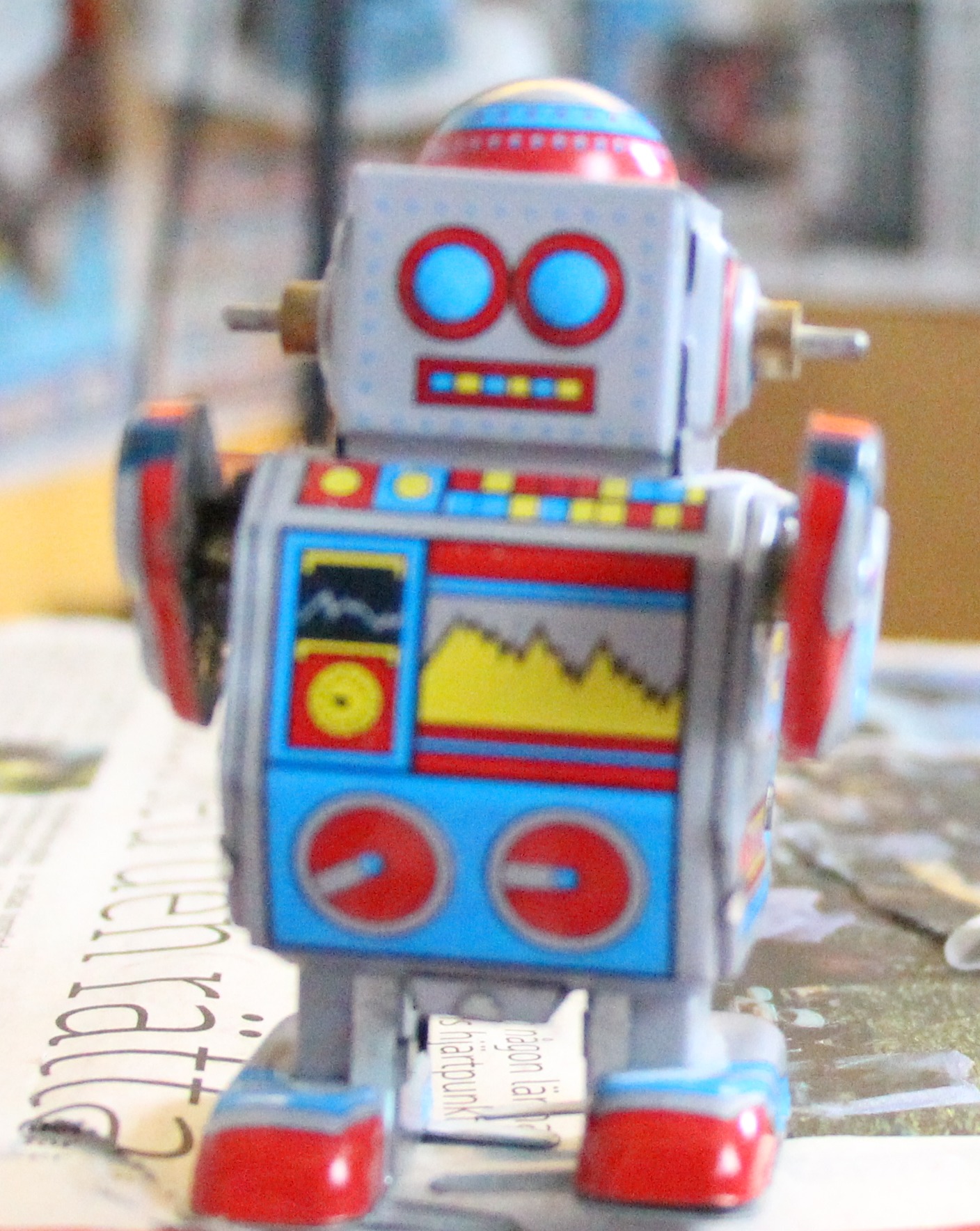} 
      \hfill
      \includegraphics[width=0.15\textwidth]{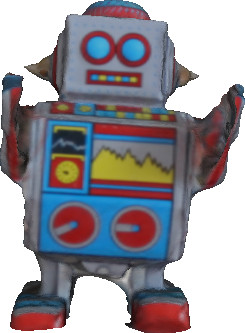}
      \hfill
      \includegraphics[width=0.14\textwidth]{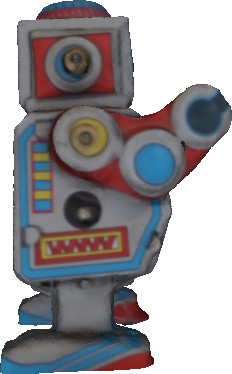}
      \hfill
      \includegraphics[width=0.14\textwidth]{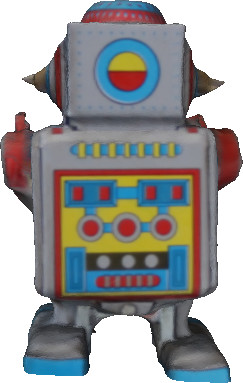}
      \hfill\null

    \null\hfill
    \subfigure[]{ \includegraphics[width=0.2\textwidth]{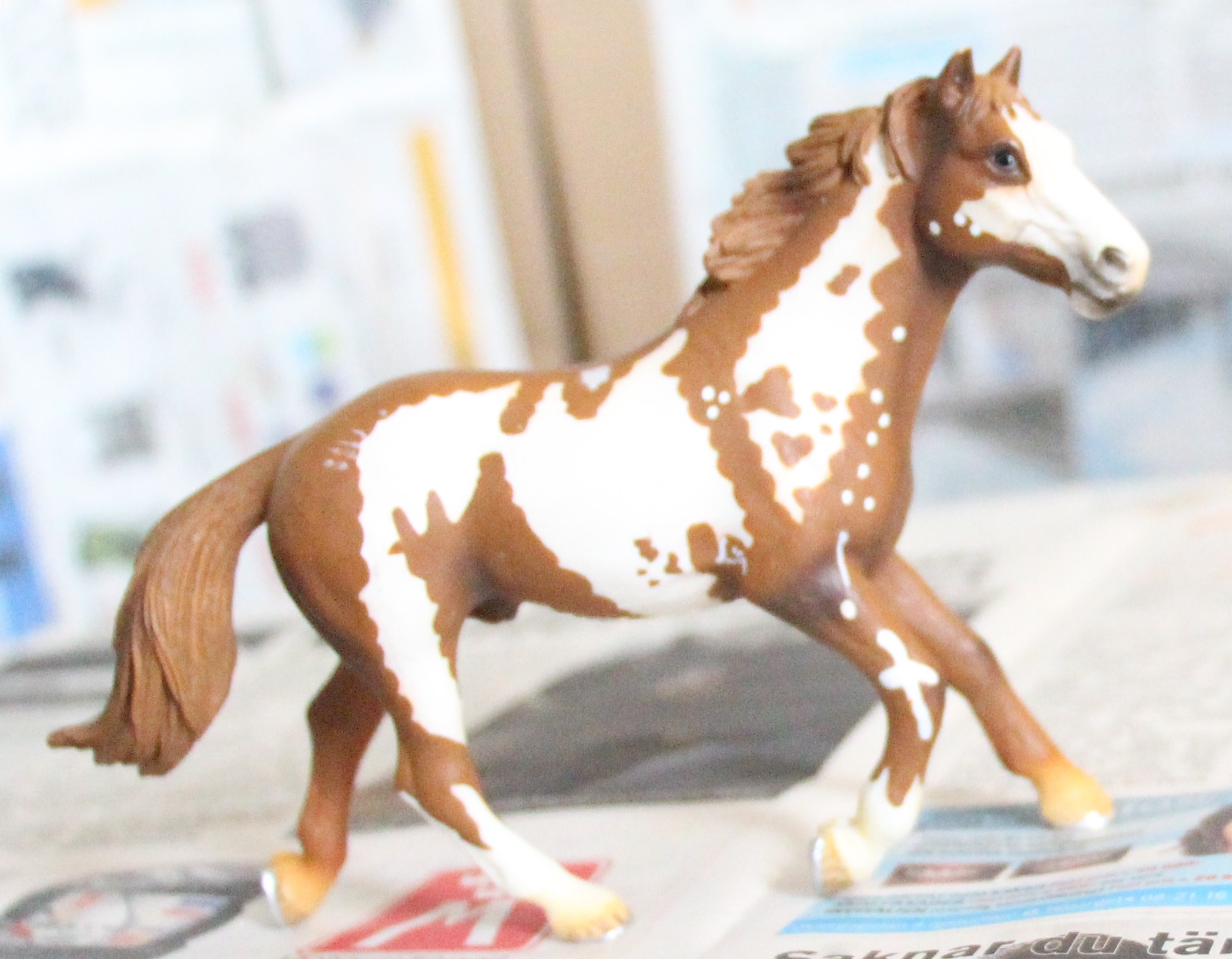}}
       \hfill
    \subfigure[]{ \includegraphics[width=0.16\textwidth]{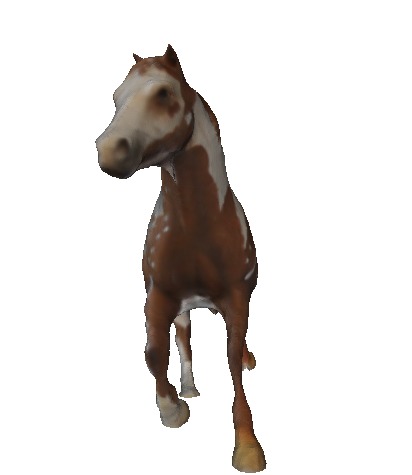}}
    \hfill
    \subfigure[]{ \includegraphics[width=0.16\textwidth]{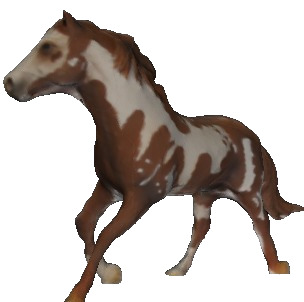}} 
    \hfill
    \subfigure[]{ \includegraphics[width=0.2\textwidth]{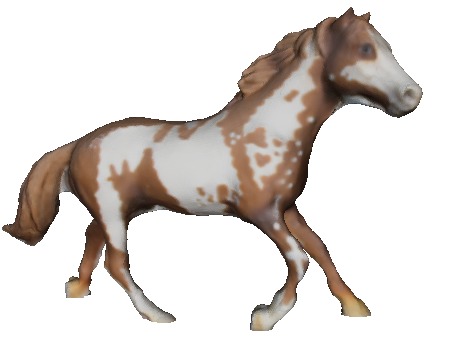}}
    \hfill\null

   \caption{Results from example objects: (a) Real images used for modeling the objects. (b)-(d) Final post-processed
   3D meshes for these objects. A cereal box (see table under category "kitchen"), a full-sized PR2 (category "other"),
   a rubber duck, a small robot toy and a horse model (category toys).}
      \label{fig:models}
\end{figure*} 
Most grasp planners \parencite{Dani, Miller, Marcus, Ken} rely on estimated object pose to parametrize grasps, i.e., to
calculate wrist position and orientation. The models obtained with the approach proposed here (see also Figure \ref{fig:models}) can
be utilized for pose estimation as can be seen in Figure \ref{fig:tracking} and Figure
\ref{fig:pr2tracking}. 
We have conducted experiments using the 3D printed objects displayed in Figure \ref{fig:models}, which are based on a
real PR2 robot, a cereal box, a toy robot, a toy horse and toy duck. Note that
these objects have complex shapes and vary in scale. The printed
objects are then tracked based on the models obtained from the original real-world objects using the tracker described in the previous section. 

We have conducted preliminary experiments with some of the object models in order to demonstrate the feasibility of
using these models for combined pose tracking and grasping purposes. For our grasping experiments, we used a robot
composed of an industrial KUKA arm and a Schunk Dexterous (SDH) Hand with a predefined hand preshape, as displayed in
Figure \ref{fig:KukaSdhExp}. Based on the estimated object pose, we executed side and top grasps by placing the wrist to
a predefined distance from the object's center along its vertical and horizontal axis and closing the fingers.
Figure \ref{fig:pr2tracking} displays tracking results based on a PR2 robot's onboard arm-camera, where the robot
detects and tracks the horse, cereal box and robot toy in the same scene while the cereal box is being lifted by
the robot. Note that the proposed tracking system can reliably handle the resulting occlusions in this scene.

\subsubsection{Replicable Manipulation Research}
Since 3D printing has become widely available, the proposed approach enables researchers to create replicable robotic experiments
by running robotic experiments using 3D printed objects. Furthermore, 3D printed objects may also serve as 
a controllable testing environment for tracking algorithms other than the proposed reference tracking system.

\subsubsection{Manipulation of Objects with Controlled Physical Properties}
The proposed 3D printing (or milling) process allows for various choices of materials such as plastics, sandstone and
wood. Using this approach, object properties can be separated from object geometry, as specified
by the scanned meshes. Furthermore, the internal mass distribution of an object can be modified by partially filling the
printed meshes, or keeping them hollow, etc. This, we believe, provides an interesting avenue for robotic manipulation
research to focus on sub-problems such as robustness to variations in friction coefficients, mass distribution, etc. 
Another important aspect is the scaling of the resulting objects. Many
current robotic hands have dimensions differing from a human adult or child's hand. By printing objects at a range of
scales, researchers may be able to study the success of robotic grasps depending on scale and could, for example, optimize
robot hand design with respect to object size. A further interesting direction of research is the study of grasps on continuous
families of perturbations of objects. A simple example would be grasping cones with various angles at the apex to
understand frictional properties, but more generally, shape and grasp \emph{moduli spaces} (\cite{pokorny2013b,
    pokorny2014a}) defined by deformations of shapes and grasps could be studied by 3D printing perturbations of existing
objects. This constitutes a direction of research we would like to investigate in future, in particular.

\section{Content and Format of Initial Core Database Release}
\label{sec:git}
The initial release of CapriDB contains 40 watertight textured mesh models of the objects listed in Table~\ref{table:objectsall} and depicted in Figure \ref{fig:objects}. 
Mesh models are stored in Wavefront OBJ format, a mesh to texture mapping is provided in MDL format and an associated texture file is stored as a JPEG image for each object. The objects for the IEEE ICRA Amazon Picking Challenge 2015 are also included in the database. Table~\ref{table:objectsall} lists the physical dimensions of these objects, their weight and original material as well as additional notes which will also be stored in CapriDB. In addition, the initial database release contains the original photos (approx. 40 per object) used to construct the mesh approximation in JPEG format. 

\begin{figure*}[t]
      \centering
     \subfigure[]{  \includegraphics[width=0.3\linewidth]{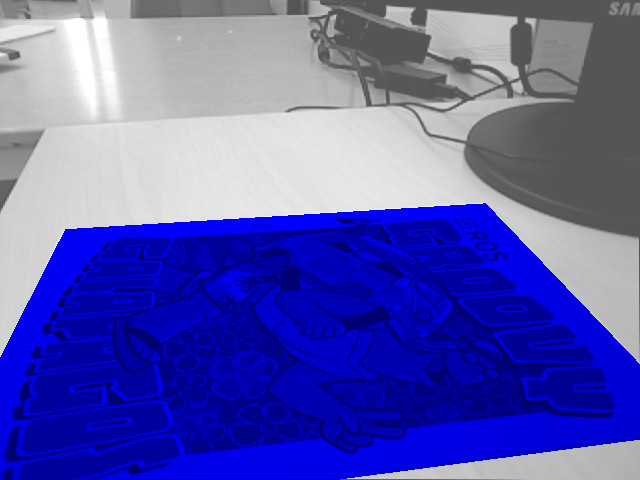}}
     \subfigure[]{  \includegraphics[width=0.3\linewidth]{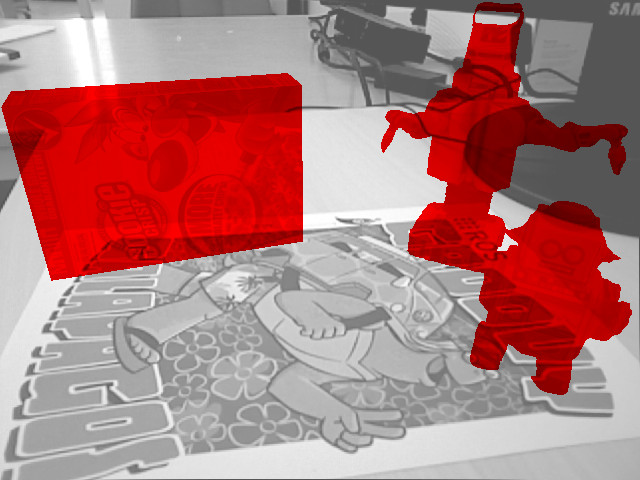}}
     \subfigure[]{  \includegraphics[width=0.3\linewidth]{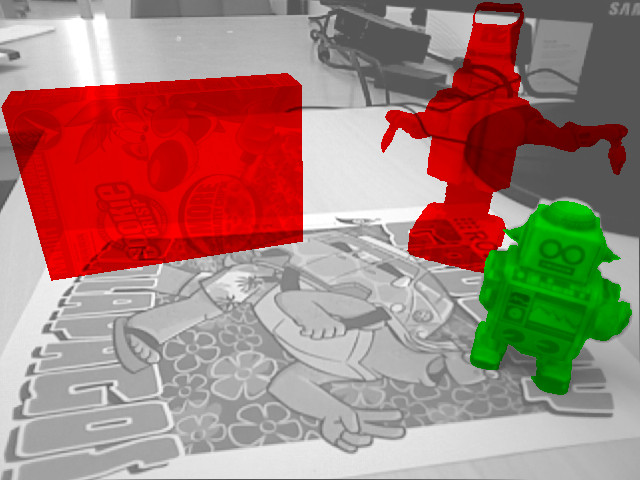}}
    
     \subfigure[]{  \includegraphics[width=0.3\linewidth]{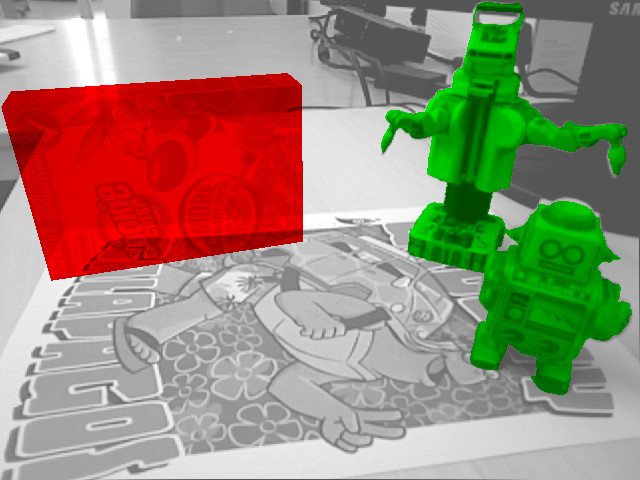}}
     \subfigure[]{  \includegraphics[width=0.3\linewidth]{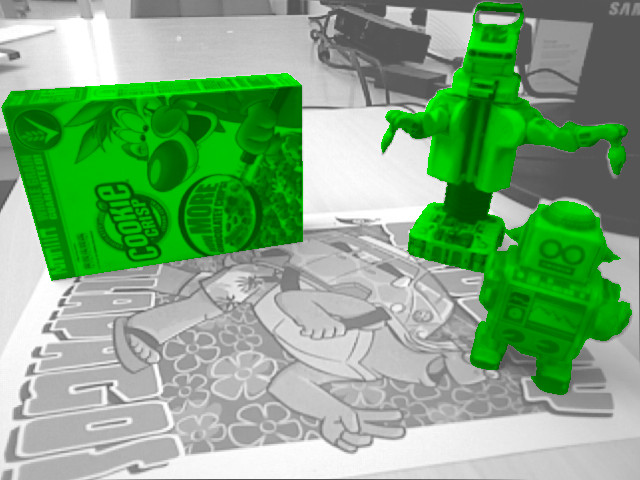}}
     \subfigure[]{  \includegraphics[width=0.3\linewidth]{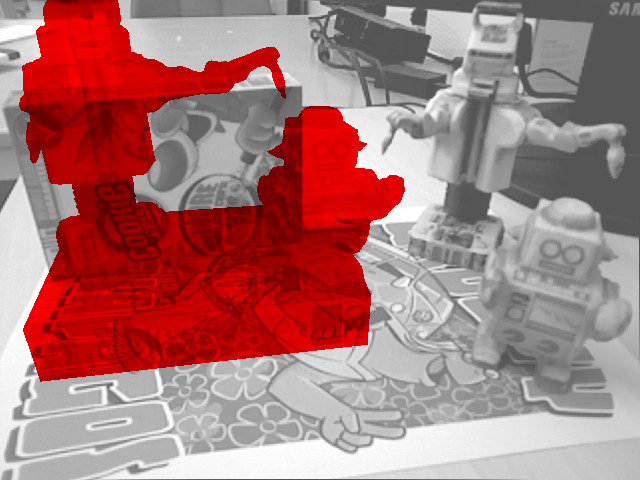}}
    
     \subfigure[]{  \includegraphics[width=0.3\linewidth]{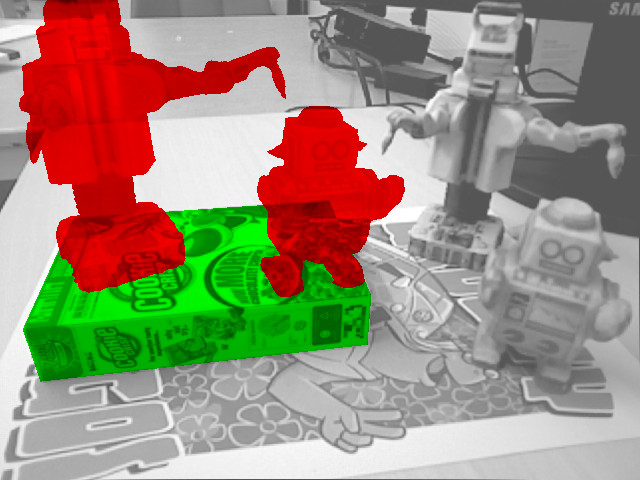}}
     \subfigure[]{  \includegraphics[width=0.3\linewidth]{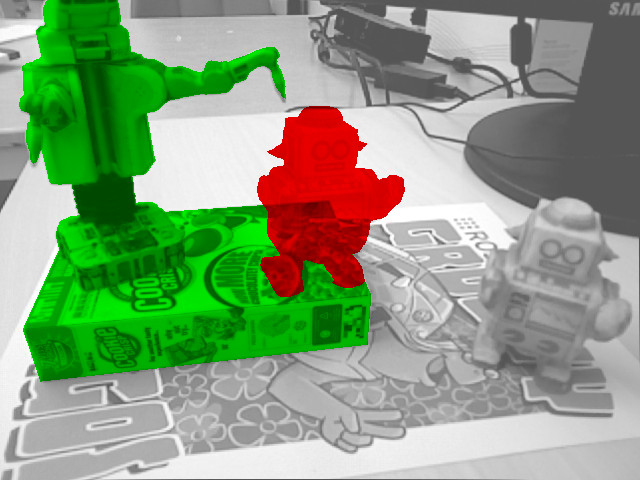}}
     \subfigure[]{  \includegraphics[width=0.3\linewidth]{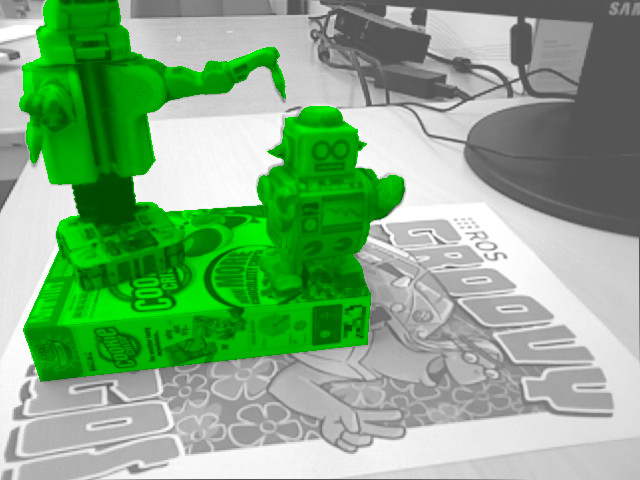}}
    
   \caption{Benchmarking using pre-defined scene layouts: (a) A marker is introduced in the scene, detected, and highlighted in blue. This marker provides a reference frame for the scene. (b) The desired object placement according to a pre-defined initial scene layout is highlighted in red. (c-e) One-by-one, the objects are placed in the scene and the color changes to green if their placement is sufficiently accurate. (f) A pre-defined target scene layout is highlighted in red. (g-i) The task is executed and the objects are moved to their target pose.
}
      \label{fig:evaluation}
\end{figure*}

To facilitate performace evaluation, we also include reference images (in JPEG) and associated tracking boundaries
(overlayed JPEG based on object poses acquired from the tracker) for each object as in Figure~\ref{fig:tracking} to test
and compare other tracking methodologies. Figure~\ref{fig:evaluation} shows how the database and interactive tracking could be used for benchmarking using a pre-defined scene layouts.
The included scenes and object poses can be used as ground truth to set up a system using these object models and the tracker. More information about the tracker's accuracy can be found in the work of \textcite{pauwels_2015}.  

\begin{table*}[h]
\label{table:objectsall}
\small
\begin{center}
\resizebox{0.88\textwidth}{!}{
 \begin{tabular}{|>{\raggedright\arraybackslash}    p{1cm}  |p{4cm} | l | l | p{3cm} | p{2cm} |}\hline
    \textbf{ID} & \textbf{Object Name} & \textbf{Dimension} & \textbf{Weight} &  \textbf{Material} &  \textbf{Other} \\ \hline
\multicolumn{6}{ |l| }{\textbf{Office}}\\ \hline 
1&  Mead Index Cards & 7.6x2x12.7 cm &  136 g & paper in plastic & Amazon \\\hline
2&Highland 6539 Self Stick Notes & 11.7x5.3x4 cm &  167.3 g & paper in plastic &  Amazon\\\hline
3&Paper Mate 12 Count Pencils  &  4.8x1.8x19.3 cm &  68 g & cardboard  &  Amazon \\\hline
4&  Elmer's Washable No-Run School Glue & 6.4x14x3 cm &  45.36 g &  plastic  & Amazon\\\hline
5& Keyboard & 2.7x47.3x18.5  &  638 g &  hard plastic  &  \\\hline
6& Scissors &  20.5x0.8x7.5 cm & 60 g  &  hard plastic, metal  &  \\\hline
7& Stapler & 5x2.7x13 cm   & 136 g  &  hard plastic  &  \\\hline
8& Hole Puncher & 5.5x9x13.9 cm   & 483 g  &  metal  &  \\\hline
9& Tipp-Ex &  7.1x2.6x2.6 cm & 60 g  &  hard plastic  &  \\\hline

 \multicolumn{6}{ |l| }{\textbf{Kitchen}}\\ \hline 
 10& First Years Take And Toss Straw Cup  &  8x8 x15 cm &  141.7 g &  plastic, paper &  Amazon\\\hline
 11&Genuine Joe Plastic Stor Sticks &  15x11.7x10.2 cm & 249.5 g &cardboard&  Amazon\\\hline
 12& Dr. Brown's Bottle Brush  &   5.3x9.7x31 cm &   39.7 g &   paper bottom with plastic & Amazon \\\hline
 13& Oreo mega stuf & 20.3x5.1x15.2 cm &  377  g & plastic  & Amazon \\\hline
 14& Cheez It Big   &  31x16.5x16.5 cm &  385.6 g &  cardboard & Amazon \\\hline
 15& Cookie Crisp Box & 29x6.8x19.3 cm & 70 g & cardboard & empty \\ \hline
 16& Plate & 1.7x20.3x20.3 cm  &  351 g      & porcelain &    \\\hline
 17& Bowl & 5.3x11.8x11.8 cm  &   80 g  & porcelain &    \\\hline
 18& Knife & 21.5x1.3x1.8cm  &  25 g  & steel &  added texture  \\\hline
 19& Fork  &  20x1.7x2.5cm  &  28 g   & steel &  added texture   \\\hline
 \multicolumn{6}{ |l| }{\textbf{Tools}}\\ \hline 
 20& Stanley Piece Precision Screwdriver Set &  19.6x9.9x2.3  cm &  99.2 g & hard plastic  & Amazon \\\hline
 21&  Hammer &  32.7x3.4x13.5 cm &  658 g  &  hard plastic, metal  &  \\\hline
 22&  Pitcher &  16.4x2.8x5.3 cm &  146 g &  hard plastic  &  \\\hline
 23&  Saw & 63.5x2.514.3 cm  &  425 g  &  hard plastic  &  \\\hline
 24&  Screwdriver & 20x2.7x2.7 cm  & 56 g  &  hard plastic  &  \\\hline
 \multicolumn{6}{ |l| }{\textbf{Toys}}\\ \hline 
 25& 6 Colored Highlighters  &  1.8x11.9x13.2 cm &  39.7 g &  plastic & Amazon \\\hline
 26& Crayola 64 Ct Crayons & 14.5x12.7 cm &  357.2 g & cardboard  & Amazon \\\hline
 27& KONG Squeakair Tennis Ball with Rope Dog Toy& 52.1x6.4x6.4 cm &  82.2 g & textile  & Amazon \\\hline
 28&  Squeakin' Eggs Plush Dog Toys & 17.8x7.6x14 cm &   8.5 g &  textile  & Amazon \\\hline
 29& KONG Squeakair Sitting Duck Dog Toy  &  12.7x5.1x8.9 cm &  8.5 g & textile, cardboard& Amazon \\\hline
 30&  KONG Squeakair Sitting Frog Dog Toy&  14x4.6x8.9 cm &  8.5 g &  textile, cardboard & Amazon \\\hline
 31& Munchkin White Hot Duck Bath Toy &  13.2x7.1x9.7 cm &  8.5 g &  textile, cardboard & Amazon \\\hline
 32& Adventures of Huckleberry Finn  &  2x13x21.6 cm &  181.44 g & paper & Amazon \\\hline
 33& Laugh-Out-Loud Jokes for Kids & 10.7x0.8x17.3 cm &    8.5 g &  paper & Amazon \\\hline

34& Black-White Duck & 11.2x7.5x11.2 cm  &  110 g   & rubber  &    \\\hline
35& Small Robot & 8.1x5.6x4.9 cm  &  60 g   & metal &    \\\hline
36& Horse & 10.3x3.2x15.7 cm  &  104  g  & hard plastic &    \\\hline
37&  Car &  4.2x5.9x12.7 cm &  145 g & metal  &    \\\hline
 \multicolumn{6}{ |l| }{\textbf{Other}}\\ \hline 
 38&  PR2 Robot &  178.9x118.6x136.9 cm &  220 kg   &  &    \\\hline
 39& Mommys Helper Outlet Plugs & 3.8x3.8x1.3 cm &   22.7 g  & plastic, cardboard & Amazon  \\\hline
 40& Expo Dry Erase Board Eraser  &  5.3x13.2x3 cm & 8.5 g & cardboard & Amazon \\\hline

   \end{tabular}
}
\end{center}
\caption{Summary of objects in our initial database release.}
\end{table*}

\section{Conclusion and Future Work}
\label{sec:conclusions}
We have introduced an inexpensive pipeline which utilizes 3D printing, inexpensive mesh reconstruction from
monocular images and a state of the art tracking algorithm to facilitate reproducible robotic manipulation research.
Our approach only requires a regular RGB or RGB-D camera and images taken from 
a set of angles and access to a 3D printing service. 
The initial database of 40 scanned objects is available at \url{http://www.csc.kth.se/capridb} and
we plan to continue contributing to this database over time by adding more objects and object features.
\section*{Acknowledgments}
This work was supported by
the EU through the project RoboHow.Cog (FP7-ICT-288533) and the Swedish Research Council.

\balance
\printbibliography

\end{document}